%% file: main.tex
\documentclass{article} 
\usepackage{iclr2022_srml_workshop,times}

\input{math_commands.tex}

\usepackage{amssymb,amsfonts,amsmath}
\usepackage{hyperref}
\usepackage{url}
\usepackage{booktabs}
\usepackage{multicol}
\usepackage{multirow}
\usepackage{subcaption}
\usepackage{graphicx}
\usepackage{algorithm}
\usepackage{algpseudocode}

\usepackage[utf8]{inputenc}
\usepackage[T1]{fontenc}
\usepackage{placeins}
\newtheorem{definition}{Definition}
\newtheorem{theorem}{Theorem}

\newcommand{\Stable}{\mathcal{X}_s}

\title{Learning Stabilizing Policies in Stochastic Control Systems}

\author {
    \DJ{}or\dj{}e \v{Z}ikeli\'c\thanks{Equal Contribution.}, 
    Mathias Lechner\footnotemark[1],
    Krishnendu Chatterjee,
    Thomas A. Henzinger \\\
    IST Austria\\\
    Klosterneuburg, Austria\\\
    \{djordje.zikelic, mathias.lechner, krishnendu.chatterjee, tah\}@ist.ac.at
}

\iclrfinalcopy 
\begin{document}

\maketitle

\begin{abstract}
In this work, we address the problem of learning provably stable neural network policies for stochastic control systems.
While recent work has demonstrated the feasibility of certifying given policies using martingale theory, the problem of how to learn such policies is little explored.
Here, we study the effectiveness of jointly learning a policy together with a martingale certificate that proves its stability using a single learning algorithm.
We observe that the joint optimization problem becomes easily stuck in local minima when starting from a randomly initialized policy. Our results suggest that some form of pre-training of the policy is required for the joint optimization to repair and verify the policy successfully.
\end{abstract}

\section{Introduction}\label{sec:intro}

Reinforcement learning (RL) has achieved impressive results in many domains in which the goal is to optimize expected reward~\citep{sutton2018reinforcement}. 
This has fueled the desire to use RL in robotics and for control of non-linear systems. However, using RL in safety-critical domains such as autonomous driving requires formal safety guarantees as consequences of unsafe behavior could be disastrous~\citep{AmodeiOSCSM16}. Verification of systems with learned components and learning with safety guarantees have thus become active research topics~\citep{GarciaF15}.

In this work, we consider the problem of learning stabilizing policies for stochastic control systems. Stability is one of the most basic and important safety properties of systems. It requires a system to be able to reach and stay within a region that is known to be safe, i.e., a stabilizing region, from any system state via the means of a stabilizing policy~\citep{lyapunov1992general}. For instance, if a self-driving car were to drive at a speed that exceeds the allowed speed limit, then a stabilizing policy would stabilize the car back within the speed limit. 

Our work builds on the recent method of~\citep{lechner2021stability} for verifying that a {\em pre-learned} control policy ensures probability $1$~stability of the system. Stability is verified by learning a {\em ranking supermartingale (RSM)}, which can be viewed as a stochastic extension of Lyapunov functions for deterministic systems~\citep{khalil2002nonlinear}. The method of~\citep{lechner2021stability} 
makes very mild assumptions on the system which make it applicable to a wide range of systems and control policies. 
However, the key drawback of the method is that it can only be used to {\em verify} probability~$1$ stability under a given policy. If the pre-learned policy does not ensure stability, the method in its current form provides no means to {\em learn} a new policy in a way that would provide stability guarantees. \citep{lechner2021stability} briefly discuss the possibility of extending the method to safe learning by jointly learning the policy and the RSM, however this idea is not empirically evaluated. 

The goal of this work is to empirically evaluate this extension in order to understand the difficulties of learning stabilizing policies as opposed to the stability verification of pre-learned policies. In particular, our results show that a good initialization of the policy is {\em essential}, as otherwise the learning algorithm easily gets stuck in a local minimum at which the learned policy cannot be verified to be ensure stability. We propose a simple-form reward function that intuitively captures the stability task and use proximal policy optimization (PPO)~\citep{schulman2017proximal} with respect to the reward function to initialize the policy. Our experimental results show that such an initialization leads to significant improvement in performance over a naive initialization of the policy.

\smallskip\noindent{\bf Related work} Learning methods present a promising approach to solving complex non-linear control tasks due to their generality. Several recent works propose learning stabilizing policies for {\em deterministic} control problems together with a Lyapunov function that certifies stability~\citep{RichardsB018,ChangRG19,AbateAGP21}. In particular, \citep{ChangRG19,AbateAGP21} propose a learner-verifier framework similar to that of~\citep{lechner2021stability} and the one presented in our work. There is a rich body of literature on theoretical stability analysis in stochastic dynamical systems and control problems, see~\citep{Kushner14} for a survey, however very few works consider automated control with stability guarantees. Control with finite-time horizon reachability constraints has been considered in~\citep{CrespoS03,SoudjaniGA15,LavaeiKSZ20,cauchi2019stochy,VinodGO19}. With the exception of~\citep{lechner2021stability} that was discussed above and that our work builds on, the work of~\citep{Vaidya15} is to our best knowledge the only work that considers control with infinite-time horizon stability constraints and that provides formal guarantees, however the stability certificates are computed in a form that makes them piecewise constant and the work verifies a weaker notion of stability called “coarse stochastic stability”.

\section{Ranking Supermartingales for Stability Verification}\label{sec:prelims}

We consider a discrete-time stochastic dynamical system defined by the equation
\[ \mathbf{x}_{t+1} = f(\mathbf{x}_t, \pi(\mathbf{x}_t), \omega_t), \]
where $f:\mathcal{X}\times\mathcal{U}\times\mathcal{N}\rightarrow\mathcal{X}$ is a dynamics function, $\pi:\mathcal{X}\rightarrow\mathcal{U}$ is a policy and $\omega_t$ is a stochastic disturbance vector sampled according to a probability distribution $d$ over $\mathcal{N}$. Here, $\mathcal{X}\subseteq\mathbb{R}^n$ is the state space of the system, $\mathcal{U}\subseteq\mathbb{R}^m$ is the action space and $\mathcal{N}\subseteq\mathbb{R}^p$ is the stochastic disturbance space, all of which are required to be Borel-measurable. We assume that $\mathcal{X}\subseteq\mathbb{R}^n$ is compact and that $f$ and $\pi$ are Lipschitz continuous, which are common assumptions in control theory.

\smallskip\noindent{\bf Almost-sure asymptotic stability} There are several notions of stability in stochastic systems, and we consider almost-sure asymptotic stability~\citep{kushner1965stability}. Let $\Stable\subseteq\mathcal{X}$ is a non-empty Borel-measurable subset that is closed under system dynamics so that the system cannot leave it once reached, i.e., for every $\mathbf{x}\in\Stable$ we have that $f(\mathbf{x},\pi(\mathbf{x}),\omega)\in\Stable$ for any $\omega\in\mathcal{N}$. We say that $\Stable$ is {\em almost-surely (a.s.) asymptotically stable}, if for each initial state $\mathbf{x}_0\in\mathcal{X}$ we have that
\[ \mathbb{P}_{\mathbf{x}_0}\Big[ \lim_{t\rightarrow\infty}\Big(\inf_{\mathbf{y}\in\Stable}||\mathbf{x}_t-\mathbf{y}||_1\Big) = 0 \Big] = 1. \]
Here, $\mathbb{P}_{\mathbf{x}_0}$ is the probability measure over the set of all system trajectories that start in the initial state $\mathbf{x}_0$ defined by the MDP semantics of the system~\citep{Puterman94}. Our definition slightly differs from that of~\citep{kushner1965stability} which considers the special case $\Stable=\{\mathbf{0}\}$. The reason is that, analogously to~\citep{lechner2021stability} and to the existing works on learning stabilizing policies in deterministic systems~\citep{BerkenkampTS017,RichardsB018,ChangRG19}, we need to assume the existence of an open neighbourhood of the origin that is known to be stable for learning to be stable, which we ensure by assuming that $\Stable$ is closed under system dynamics. 

\smallskip\noindent{\bf Ranking supermartingales} Intuitively, a ranking supermartingale for the set $\Stable$ is a nonnegative continuous function $V:\mathcal{X}\rightarrow\mathbb{R}$ which strictly decreases in expected value upon every one-step execution of the system, until the set $\Stable$ is reached. The name comes from its connection to supermartingale processes in probability theory~\citep{Williams91}.

\begin{definition}\label{def:ranksm}
A {\em ranking supermartingale (RSM)} for $\Stable$ is a nonnegative continuous function $V:\mathcal{X}\rightarrow\mathbb{R}$ for which there exists $\epsilon>0$ such that, for every $\mathbf{x}\in\mathcal{X}\backslash \Stable$, we have that
\[ \mathbb{E}_{\omega\sim d} [ V(\mathbf{x},\pi(\mathbf{x}),\omega) ] \leq V(\mathbf{x}) - \epsilon. \]
\end{definition}

The following theorem is the key result on the use of RSMs for a.s.~asymptotic stability analysis.

\begin{theorem}[\cite{lechner2021stability}]
Let $f:\mathcal{X}\times\mathcal{U} \times \mathcal{N}\rightarrow \mathcal{X}$ be a Lipschitz continuous dynamics function, $\pi:\mathcal{X}\rightarrow\mathcal{U}$ a Lipschitz continuous policy and $d$ a distribution over $\mathcal{N}$. Suppose that $\mathcal{X}$ is compact and let $\Stable\subseteq\mathcal{X}$ be closed under system dynamics and have a non-empty interior. Suppose that there exists an RSM $V:\mathcal{X}\rightarrow\mathbb{R}$ for $\Stable$. Then $\Stable$ is a.s.~asymptotically stable.
\end{theorem}

\section{Algorithm for Learning Ranking Supermartingales}\label{sec:algo}

We now present the details behind our learner-verifier framework for simultaneously learning a stabilizing policy $\pi_\mu$ and an RSM $V_\theta$ for $\Stable$, both of which are parametrized as neural networks with parameters $\mu$ and $\theta$. The verifier module is analogous to that of~\citep{lechner2021stability} for verifying a learned RSM candidate under the fixed policy, so we only provide a high level presentation of it. The novelty of our algorithm lies in modifying the learner in a way which allows learning $\pi_\mu$ and $V_\theta$ simultaneously as briefly discussed in~\citep{lechner2021stability}, and in observing the importance of algorithm initialization. The algorithm pseudocode is provided in Appendix~\ref{app:pseudocode}.

\smallskip\noindent{\bf Initialization} Our algorithm first initializes the policy $\pi_\mu$ by using proximal policy optimization (PPO)~\citep{schulman2017proximal}. In particular, the algorithm first induces an MDP from the system and defines a reward function $r:\mathcal{X}\rightarrow [0,1]$ via $r(\mathbf{x}) := \mathbb{I}[\Stable](\mathbf{x})$. The first learner-verifier iteration is then started by a call to the verifier, followed by a call to the learner. This contrasts the verification algorithm of~\citep{lechner2021stability}, which starts with the call to the learner. Our experimental results will show a naive initialization often does not allow learning a stabilizing policy. The importance of initialization was also observed in~\citep{ChangRG19} which showed that a proper initialization of the policy is necessary for a similar method for stability verification of deterministic neural network-controlled systems. Specifically, \citep{ChangRG19} initialized their linear policy with the LQR solution of the system linearized at the origin. However, it is not clear how one could linearize stochastic dynamics with non-additive kinds of stochastic disturbance or systems that are highly non-linear, whereas our initialization is applicable to general classes of stochastic disturbances and system dynamics.

\smallskip\noindent{\bf Verifier} Due to the continuity of $V_\theta$ and the compactness of $\mathcal{X}$, $V_\theta$ is bounded from below and therefore may be increased by a constant value in order to make it nonnegative. Hence, the verifier only needs to check whether the RSM candidate satisfies the expected decrease condition. To do this, the algorithm computes a discretization the part of the state space $\mathcal{X}\backslash\Stable$. A {\em discretization} of $\mathcal{X}\backslash\Stable$ with {\em mesh} $\tau>0$ is a set $\tilde{\mathcal{X}}\subseteq\mathcal{X}$ such that, for every $\mathbf{x}\in \mathcal{X}\backslash\Stable$, there exists $\tilde{\mathbf{x}}\in\tilde{\mathcal{X}}$ such that $||\mathbf{x}-\tilde{\mathbf{x}}||_1<\tau$. 
As $f$ is assumed to be Lipschitz continuous and as $\pi_\mu$ and $V_\theta$ are neural networks with continuous activation functions thus also Lipschitz continuous~\citep{SzegedyZSBEGF13}, we can show that the expected decrease condition may be checked by checking a slightly stricter condition at the discretization points. In particular, let $L_V$, $L_f$ and $L_\pi$ being Lipschitz constants of $V_\theta$, $f$ and $\pi_\mu$. We assume that $L_f$ is provided, and use the method of~\citep{SzegedyZSBEGF13} to compute $L_V$ and $L_\pi$. Let $K = L_V \cdot (L_f \cdot (L_\pi + 1) + 1)$. Then, if we show that for every $\tilde{\mathbf{x}}\in\tilde{\mathcal{X}}$ we have
\begin{equation}\label{eq:lipschitz}
    \mathbb{E}_{\omega\sim d}\Big[ V_\theta \Big( f(\tilde{\mathbf{x}}, \pi_\mu(\tilde{\mathbf{x}}), \omega) \Big) \Big] < V_\theta(\tilde{\mathbf{x}}) - \tau \cdot K,
\end{equation}
then there exists $\epsilon>0$ for which $V_\theta$ satisfied the expected decrease condition~\citep[Theorem~3]{lechner2021stability}. Hence, the verifier may check that $V_\theta$ is an RSM by simply checking whether the condition in eq.~(\ref{eq:lipschitz}) is satisfied at each $\tilde{\mathbf{x}}\in\tilde{\mathcal{X}}$. The expected value on the left hand side of the above inequality is computed by using interval arithmetic abstract interpretation, see~\citep{lechner2021stability} for details. If the verifier concludes that eq.~(\ref{eq:lipschitz}) is satisfied at each $\tilde{\mathbf{x}}\in\tilde{\mathcal{X}}$, the algorithm concludes a.s.~asymptotic stability of $\Stable$. Otherwise, for each state $\tilde{\mathbf{x}}\in\tilde{\mathcal{X}}$ at which the condition is violated, the verifier samples $N$ successor states of $\tilde{\mathbf{x}}$ where $N$ is an algorithm parameter, and adds them to a set $\mathcal{D}_{\tilde{\mathbf{x}}}$ that will be used by the learner to approximate the expected value $\mathbb{E}_{\omega\sim d}[ V_\theta( f(\tilde{\mathbf{x}}, \pi_\mu(\tilde{\mathbf{x}}), \omega))]$. 

\smallskip\noindent{\bf Learner} A policy $\pi_\mu$ and an RSM candidate $V_\theta$ are learned by minimizing the loss function
\begin{equation}\label{eq:loss}
    \mathcal{L}(\mu, \theta) = \mathcal{L}_{\text{RSM}}(\mu, \theta) + \lambda \cdot  \mathcal{L}_{\text{Lipschitz}}(\mu, \theta).
\end{equation}
The first loss term $\mathcal{L}_{\text{RSM}}(\mu, \theta)$ is defined via
\begin{equation*}
    \mathcal{L}_{\text{RSM}}(\mu, \theta) = \frac{1}{|\tilde{\mathcal{X}}|}\sum_{\tilde{\mathbf{x}}\in\tilde{\mathcal{X}}}\Big( \max\Big\{\sum_{\mathbf{x}'\in\mathcal{D}_{\tilde{\mathbf{x}}}}\frac{V_{\theta}(\mathbf{x}')}{|\mathcal{D}_{\tilde{\mathbf{x}}}|}-V_{\theta}(\tilde{\mathbf{x}}) + \tau \cdot K, 0\Big\} \Big).
\end{equation*}
Intuitively, each term in the sum is used to guide the learner towards learning an RSM candidate that satisfies the expected dicrease condition at the state $\tilde{\mathbf{x}}\in \mathcal{X}$. Since we cannot compute the closed form expression for the expected value of $V_{\theta}$ over the successor states of $\tilde{\mathbf{x}}$, we use sampled successor states that were computed by the verifier at counterexample states. Such an on-demand sampling of successor states improves the scalability of our algorithm. The second loss term $\lambda\cdot \mathcal{L}_{\text{Lipschitz}}(\mu, \theta)$ in the sum is a regularization term used to guide the learner towards learning $\pi_\mu$ and $V_\theta$ with small Lipschitz constant, so as to make the term $\tau\cdot K$ in the condition checked by the verifier as small as possible. 
We defer the details on the regularization term to Appendix~\ref{app:loss}.

\section{Experimental Evaluation}\label{sec:experiments}


This section investigates the importance of policy initialization.

First, we consider the performance of both the algorithm of~\citep{lechner2021stability} which keeps the policy's parameters $\mu$ frozen and its extension to an algorithm that learns a stabilizing policy by making $\mu$ trainable. We repeat this experiment for a different number of policy pre-training iterations with PPO and random seeds ($n$=10). We then report the algorithm's success, i.e., whether a valid RSM is found.
We use 2D system benchmark environment of \citet{lechner2021stability} for our experimental study. Details on the experiment setup can be found in Appendix \ref{app:training} and \ref{app:ppo}.

\begin{table}[t]
\vspace{-0.2cm}
    \centering
    \caption{Number of certified stable instances with different number of PPO pre-training iterations.}
    \vspace{-0.2cm}
    \begin{tabular}{c|c|c}\toprule
             Number of PPO iterations  &\ \ \  $\mu$ fixed\ \ \  & $\mu$ learnable \\\midrule
         0 & 0/0 & 0/0 \\
         20 & 0/10 & 1/10  \\
         30 &  6/10 & 7/10 \\
         40 & 9/10 & 10/10 \\
         50 & 10/10 & 10/10 \\\bottomrule
    \end{tabular}
    \label{tab:results}
    \vspace{-0.2cm}
\end{table}

The results in Table \ref{tab:results} show that, without a proper initialization, the algorithm cannot learn a safe policy and a valid RSM regardless of whether the policy's parameters $\mu$ are fixed or trainable. After a short pre-training of 20 PPO iterations, the algorithm could only learn a stable policy in 1 out of 10 trials.
Conversely, if the initially provided policy is well-trained, the algorithm can prove its stability even if the policies' parameters are frozen for all ten tested runs. 
Aggregated over all 40 runs, only in 3 cases (7.5\%) the joint training of the policy and the RSM helped over just training an RSM with a fixed pre-trained policy.

\begin{figure}[t]
\vspace{-0.1cm}
\centering
\subcaptionbox{20 PPO pre-training iterations (failed)\label{fig:suba}}
{\includegraphics[width=0.45\textwidth]{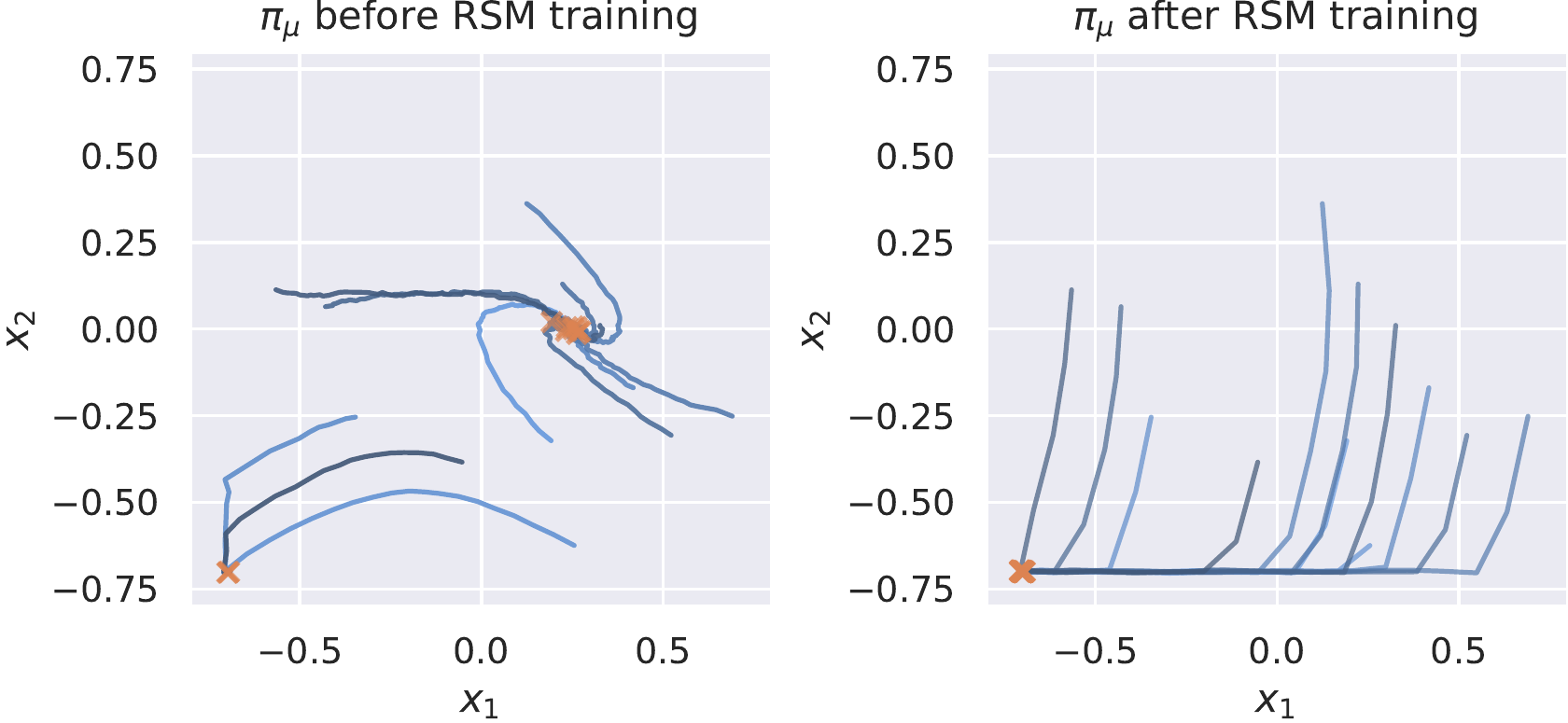}}
\subcaptionbox{30 PPO pre-training iterations (success)\label{fig:subb}}
{\includegraphics[width=0.45\textwidth]{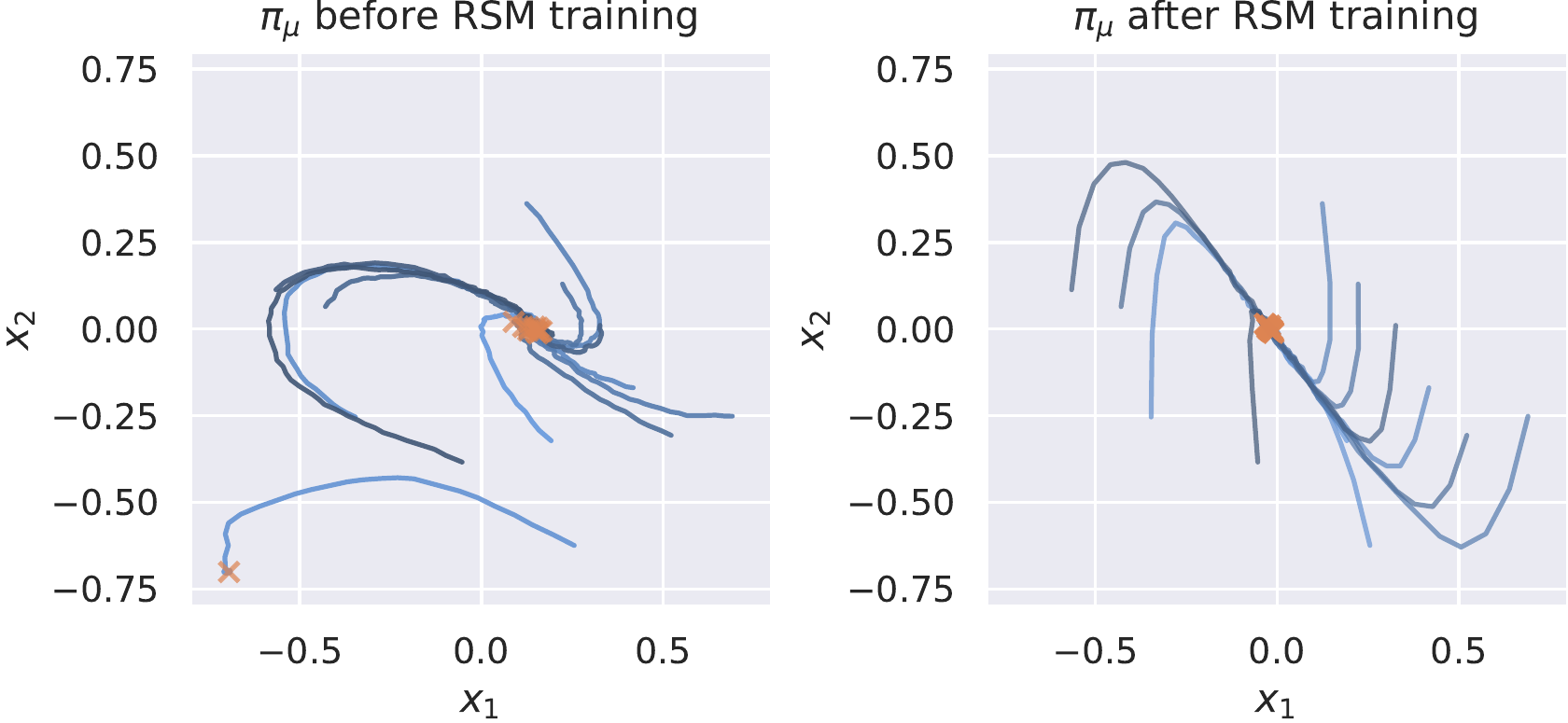}}
\caption{System trajectories before and after RSM learning with Algorithm \ref{alg:algorithm}.  The orange markers indicate the trajectories' terminal states after 200 steps. The policy is pre-trained with 20 (\ref{fig:suba}) and 30 (\ref{fig:subb}) PPO iterations. The RSM learning process in \ref{fig:suba} is stuck in a local minimum where the RSM decrease condition is satisfied in all except one state (left bottom). The RSM learning process in \ref{fig:subb} terminated successfully, proving that policy stabilizes the system to a set around the origin.}
\label{fig:localmin}
\vspace{-0.3cm}
\end{figure}

In Figure \ref{fig:localmin}, we analyze the difficulties of the joint optimization problem. Figure \ref{fig:suba} shows that the training algorithm gets stuck in a local minimum without a good initialization. The expected decrease condition is fulfilled everywhere on the state space except in a single point (bottom left). Consequently, the system in Figure \ref{fig:suba} experiences a low training loss but our algorithm learns an invalid RSM. Escaping this sub-optimal minimum requires the RSM and the policy to change their behavior on the entire state space, making it improbable using stochastic gradient descent.
Figure \ref{fig:subb} visualizes an example when the pre-trained policy is unsafe but good enough for our algorithm to repair it to a certifiable safe policy.

\smallskip\noindent{\bf Acknowledgements} This work was supported in part by the ERC-2020-AdG 101020093, ERC CoG 863818 (FoRM-SMArt) and the European Union’s Horizon 2020 research and innovation programme under the Marie Skłodowska-Curie Grant Agreement No.~665385.

\bibliography{bibliography}
\bibliographystyle{iclr2022_conference}

\newpage
\appendix
\section{Appendix}

\subsection{Algorithm Pseudocode}\label{app:pseudocode}

\begin{algorithm}
\caption{Learning a stabilizing policy}
\label{alg:algorithm}
\begin{algorithmic}[1]
\State \textbf{Input} Dynamics function $f$, disturbance distribution $d$, region $\Stable\subseteq\mathcal{X}$ \\
\hspace{0.9cm} Lipschitz constant $L_f$, parameters $\tau>0$, $N\in\mathbb{N}$, $\lambda > 0$

\State $\pi_\mu \leftarrow$ policy initialized via PPO 
\State $\tilde{\mathcal{X}} \leftarrow $ discretization of $\mathcal{X}\backslash \Stable$ with mesh $\tau$
\For{$\mathbf{x}$ in $\tilde{\mathcal{X}}$}
\State $\mathcal{D}_{\mathbf{x}}\leftarrow$ $N$ sampled successor states of $\mathbf{x}$
\EndFor
\State $V_\theta \leftarrow$ trained RSM candidate by minimizing the loss in eq.~(\ref{eq:loss}) with parameters $\mu$ fixed
\While{timeout not reached}
\State $L_\pi$, $L_V \leftarrow$ Lipschitz constants of $\pi_\mu$, $V_\theta$
\State $K\leftarrow L_V \cdot (L_f \cdot (L_\pi + 1) + 1)$
\If{$\exists \mathbf{x}\in\tilde{\mathcal{X}}$ s.t. $\mathbb{E}_{\omega\sim d}[ V ( f(\mathbf{x}, \pi(\mathbf{x}), \omega) ) ] \geq V(\mathbf{x}) - \tau \cdot K$}
\State $\mathcal{D}_{\mathbf{x}}\leftarrow$ add $N$ sampled successor states of $\mathbf{x}$
\State $\pi_{\mu}, V_\theta \leftarrow$ trained policy and RSM candidate by minimizing the loss in eq.~(\ref{eq:loss})
\Else
\State \textbf{Return} A.s.~asymptotically stable
\EndIf
\EndWhile
\State \textbf{Return} Unknown
\end{algorithmic}
\end{algorithm}

\subsection{Lipschitz Loss Term}\label{app:loss}
The loss term $\lambda \cdot  \mathcal{L}_{\text{Lipschitz}}(\mu, \theta)$ is the regularization term used to guide the learner towards learning a policy $\pi_\mu$ and an RSM candidate $V_\theta$ such that the Lipschitz constant $L_\theta$ does not exceed a tolerable threshold $\delta>0$. This is done in order to make the term $\tau\cdot K$ in the condition verified by the verifier sufficiently small. The constant $\lambda>0$ is an algorithm parameter balancing the two loss terms, and we define
\[ \mathcal{L}_{\text{Lipschitz}}(\mu, \theta) = \max\Big\{L_V - \frac{\delta}{\tau \cdot (L_f \cdot (L_\pi + 1) + 1)}, 0 \Big\}, \]
where the Lipschitz constants $L_\pi$ and $L_\theta$ are computed as in~\citep{SzegedyZSBEGF13}.

\subsection{Training Details}\label{app:training}
The networks size and training hyperparameters for all network involved in our experiments are listed in Table \ref{tab:networks}. 
\begin{table}[]
    \centering
    \begin{tabular}{c|cccc}\toprule
         Network & Hidden size & Learning rate & Lipschitz threshold & Output activation  \\\midrule
         Policy $\pi_\mu$ & 128, 128 & $5\cdot 10^{-5}$ & 3.0 & -\\
          RSM $V_\theta$ & 128, 128 & $5\cdot 10^{-4}$ & 8.0 & Softplus \\
         Value function (for PPO) & 128, 128 & $5\cdot 10^{-4}$ & - &  - \\\bottomrule
    \end{tabular}
    \caption{Hyperparameters of the networks used in our experiments. All networks use a ReLU activation function in their hidden layers. The regularization factor $\lambda$ was set to $10^{-3}$.}
    \label{tab:networks}
\end{table}

\subsection{PPO Details}\label{app:ppo}
The settings used for the PPO \cite{schulman2017proximal} pre-training process are as follows.
In each PPO iteration, 30 episodes of the environment are collected in a training buffer. Stochastic is introduced to the sampling of the policy network $\pi_\mu$ using a Gaussian distributed random variable added to the policy's output, i.e., the policy predicts a Gaussian's mean. 
The standard deviation of the Gaussian is dynamic during the policy training process according to a linear decay starting from 0.5 at first PPO iteration to 0.05 at PPO iteration 50. The advantage values are normalized by subtracting the mean and scaling by the inverse of the standard deviation of the advantage values of the training buffer. The PPO clipping value $\varepsilon$ is 0.2 and $\gamma$ is set to 0.99.
In each PPO iteration, we train the policy for 10 epochs, except for the first iteration where we train the policy for 30 epochs. An epoch accounts to a pass over the entire data in the training buffer, i.e., the data from the the rollout episodes. 
We train the value network 5 epochs, expect in the first PPO iteration, where we train the value network for 10 epochs. The Lipschitz regularization is applied to the learning of the policy parameters during the PPO pre-training.

\end{document}

%% file: math_commands.tex
\usepackage{amsmath,amsfonts,bm}

\def\eqref#1{equation~\ref{#1}}

\def\1{\bm{1}}

\DeclareMathAlphabet{\mathsfit}{\encodingdefault}{\sfdefault}{m}{sl}
\SetMathAlphabet{\mathsfit}{bold}{\encodingdefault}{\sfdefault}{bx}{n}

